\documentclass[final]{cvpr}
\usepackage{mmstyle}
\usepackage{times}
\usepackage{epsfig}
\usepackage{graphicx}
\usepackage{amsmath}
\usepackage{amssymb}
\usepackage{algorithm}
\usepackage{algorithmicx}
\usepackage{algpseudocode}
\usepackage{pifont}
\usepackage{comment}

\usepackage{caption}
\usepackage{booktabs}  
\usepackage{threeparttable}  
\usepackage{multicol}  
\usepackage{multirow}  
\usepackage{tabu} 
\usepackage{bm}
\usepackage{float}
\usepackage[normalem]{ulem}
\usepackage[british,UKenglish,USenglish,english,american]{babel}

\newlength\savewidth\newcommand\shline{\noalign{\global\savewidth\arrayrulewidth
  \global\arrayrulewidth 1pt}\hline\noalign{\global\arrayrulewidth\savewidth}}

\newcommand{\Tref}[1]{Table~\ref{#1}}

\newcommand{\Fref}[1]{Figure~\ref{#1}}

\usepackage[pagebackref=true,breaklinks=true,colorlinks,bookmarks=false]{hyperref}

\def\cvprPaperID{3153} 
\def\confYear{CVPR 2021}

\begin{document}

\title{Are Fewer Labels Possible for Few-shot Learning?}

\author{Suichan Li$^{1,}$\thanks{Equal contribution, $\dagger$ Dongdong Chen is the corresponding author}, Dongdong Chen$^{2,*,\dagger}$, Yinpeng Chen$^{2}$, Lu Yuan$^{2}$, Lei Zhang$^{2}$, Qi Chu$^{1}$, Nenghai Yu$^{1}$, \\
$^{1}$University of Science and Technology of China 
\quad\quad $^{2}$Microsoft Research\\
\{lsc1230@mail., qchu@, ynh@\}ustc.edu.cn, cddlyf@gmail.com, \\
\{ yiche, luyuan, leizhang\}@microsoft.com}

\maketitle

\begin{abstract}
Few-shot learning is challenging due to its very limited data and labels. Recent studies in big transfer (BiT) show that few-shot learning can greatly benefit from pretraining on large scale labeled dataset in a different domain. This paper asks a more challenging question:
``can we use as few as possible labels for few-shot learning in both pretraining (with no labels) and fine-tuning (with fewer labels)?''. 

Our key insight is that the clustering of target samples in the feature space is all we need for few-shot finetuning. It explains why the vanilla unsupervised pretraining (poor clustering) is worse than the supervised one. In this paper, we propose transductive unsupervised pretraining that achieves a better clustering by involving target data even though its amount is very limited. The improved clustering result is of great value for identifying the most representative samples (``eigen-samples'') for users to label, and in return, continued finetuning with the labeled eigen-samples further improves the clustering. 
Thus, we propose eigen-finetuning to enable \textbf{fewer} shot learning by leveraging the co-evolution of clustering and eigen-samples in the finetuning. We conduct experiments on 10 different few-shot target datasets, and our average few-shot performance outperforms both vanilla inductive unsupervised transfer and supervised transfer by a large margin. For instance, when each target category only has 10 labeled samples, the mean accuracy gain over the above two baselines is \textbf{9.2\%} and \textbf{3.42\%} respectively.

\end{abstract}

\section{Introduction}
Few-shot learning~\cite{garcia2017few,ren2018meta,chen2019closer,wang2020generalizing} is a challenging problem since a very small amount of data and labels are available for training It is practically useful because accruing enough task-specific data with supervised information is prohibitively expensive. Transfer learning offers a solution: direct training on such limited data and labels is replaced with a pretraining phase. The recent work BiT~\cite{chen2020big} has shown that few-shot learning can significantly benefit from model pretraining with auxiliary large-scale labeled data (\eg, ImageNet-1k~\cite{deng2009imagenet}, ImageNet-22k~\cite{deng2009imagenet}) in a different domain. However, such big datasets with fully supervised information can be laborious to collect.

In this paper, we explore the minimum amount of labels in the few-shot learning to answer: ``\textit{is it possible to use {\textbf{\emph{fewer}}} labels for few-shot learning}?''. The question consists of two perspectives: 
\begin{itemize}
    \item Is it possible to use \textit{unlabeled} auxiliary data for pretraining?
    \item Can we further \textit{reduce the amount of labels} in task-specific data for fine-tuning?
\end{itemize}

\begin{figure}[t]
\centering
	\includegraphics[width=0.9\linewidth]{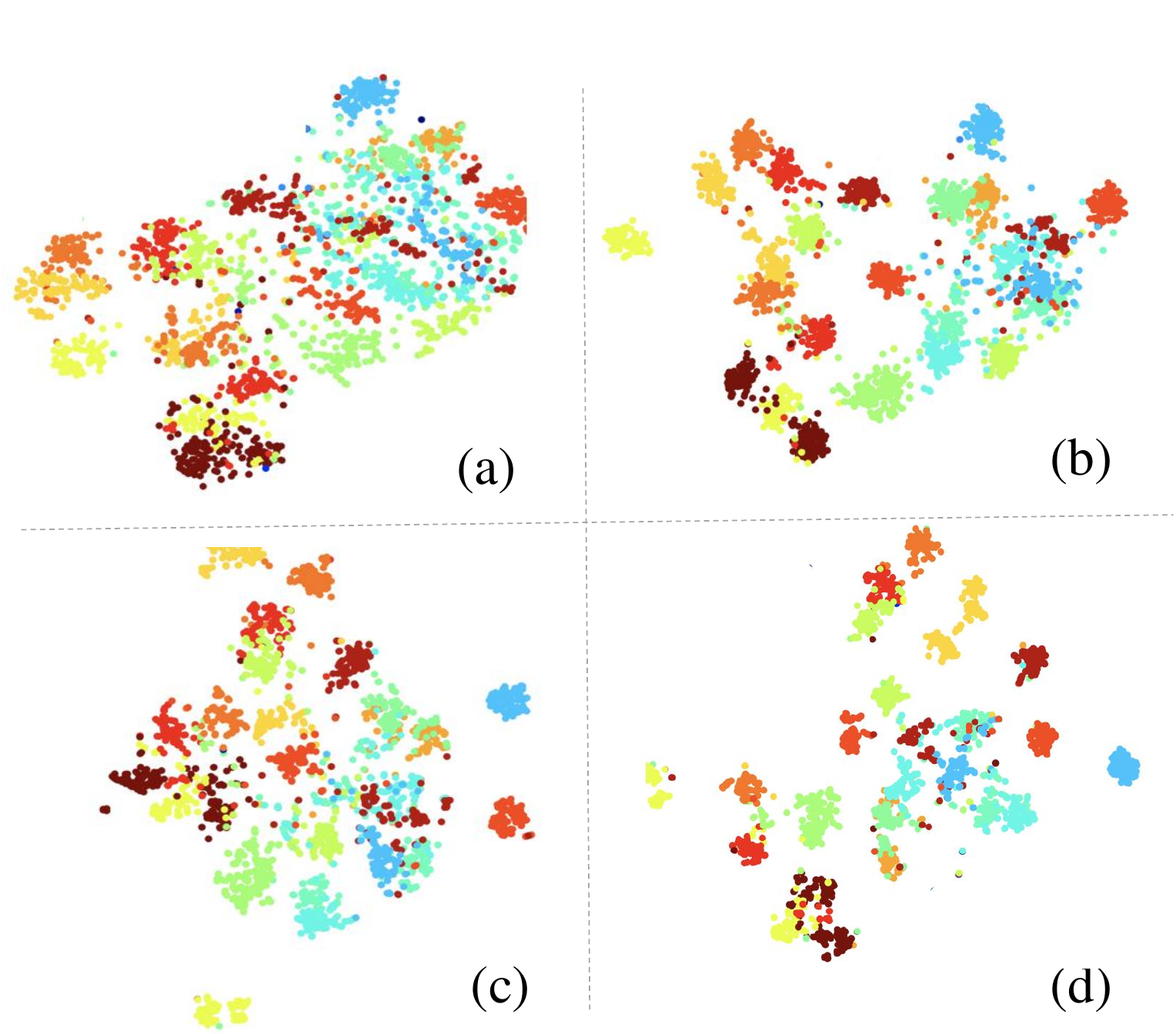}
	\caption{\small t-SNE visualization of features on Pet~\cite{parkhi2012cats} by using different models: (a) vanilla unsupervisedly pretrained model, (b)supervisedly pretrained model, (c) the proposed transductive unsupervisedly pretrained model, (d) finetuned transductive model by using a few labeled samples. } \label{fig:case}
\end{figure}

\noindent A direct solution for the first question is vanilla unsupervised transfer \cite{he2020momentum}, \ie, pretraining on a large-scale auxiliary data (source) with absence of labels and then finetuing on the task-specific data (target) with few labels. However, the un-supervised pretraining does not work as well as supervised pretraining (\ie,BiT~\cite{chen2020big}) in transfer. We compare un-supervised and supervised pretrainings in terms of the clustering of target samples in the feature space (see in \Fref{fig:case} (a) (b)), and observe the target samples are better clustered in supervised pretraining. Hence, our hypothesis is that the clustering quality of target samples matters to few-shot transfer.

More importantly, we find that the target samples have much better clustering (see \Fref{fig:case} (c)) when involving them into the unsupervised pretraining (by sample re-balancing), even though their amount is very limited. This results in a significant improvement in few-shot learning over the vanilla unsupervised pretraining. We name it \emph{transductive un-supervised pretraining} due to the involvement of the target data, while the vanilla unsupervised pretraining is ``inductive'' on the source data alone.

To address the second sub-question, we propose a novel \emph{eigen-finetuning} approach to make even \textbf{\emph{fewer}} shot learning possible. Our motivation is that the improved clustering from \emph{transductive un-supervised pretraining} can help identify the most representative samples, called \emph{``eigen-samples''}, to annotate; and finetuning the pretrained model with the labeled \emph{eigen-samples} can continue to improve clustering, as shown \Fref{fig:case} (d). By leveraging the co-evolution of clustering and eigen-samples in the finetuning, the learning is most likely to utilize fewer representative samples to annotate. Such an active learning scheme tends to be more reasonable in practice, compared to the existing ideal few-shot setting, which assumes all the labels are pre-known and randomly chooses labeled samples in a class-balance way. Consuming labels beforehand is usually unrealistic and random sampling may not be the most representative.

Equipped with the above two key designs, we significantly improve the few-shot learning performance under unsupervised pretraining and make it possible to outperform supervised pretraining. On 10 different target datasets, our average few-shot performance outperforms both vanilla unsupervised few-shot transfer and supervised few-shot transfer by a large margin. Specifically, when each target category only has 10 labeled samples, we beats the vanilla unsupervised transfer by $\bf{9.2\%}$ and supervised transfer by $\bf{3.42\%}$ respectively.  Our findings include: 

\begin{itemize}
 \item The clustering of target samples in the feature space is all we need for few-shot transfer. The clustering is significantly improved both in \emph{transductive un-supervised pretraining} and in \emph{eigen-sample finetuning}, making the few-shot performance competitive to that from supervised pretraining.
 
 \item The proposed transfer strategy utilizes as few labeled samples as possible for few-shot learning. It not only relieves the burden of collecting large-scale supervised data for pretraining, but also allows less annotation in task-specific data. 
 
 \item The proposed idea can always bring significant gains over the vanilla unsupervised pretraining across different target datasets. If the target dataset has a relatively large amount of unlabeled samples, the advantage of our approach is more significant. 
\end{itemize}

\begin{figure*}[t]
	\centering
	\includegraphics[width=1.0\linewidth]{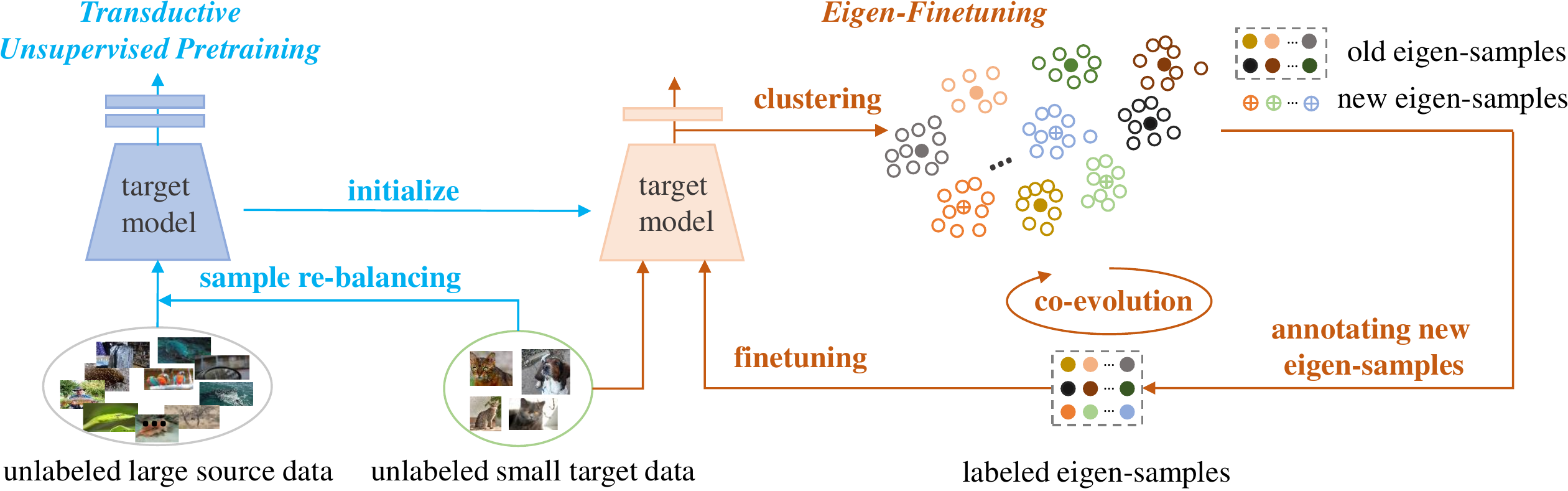}
	\caption{The proposed unsupervised pretraining framework for few-shot learning, which has two key components: transductive unsupervised pretraining and eigen-finetuing. By involving target data into pretraining, tranductive unsupervised pretraining can get better clustering in the target space. Eigen-finetuning co-evolves the process of eigen-sample selecting by clustering and model-finetuning.} 
	\label{fig:pipeline}
\end{figure*}

\section{Related works}
\paragraph{Model Pretraining.} Model pretraining plays a key role in deep learning literature. By pretraining the model on a large auxiliary source dataset and then fine-tuning on the target dataset, it can achieve better performance than the train-from-scratch counterpart, especially when the target data are very limited scale. Recently, BiT~\cite{kolesnikov2019big} showed the big success of large scale supervised pretraining on downstream few-shot learning. Meanwhile, SimCLRv2\cite{chen2020big} explored self-supervised pretraining on the large scale unlabeled dataset and showed big self-supervised model can server as a strong semi-supervised learner. In this work, we also study self-supervised pretraining on large unlabeled data. Different from \cite{chen2020big}, our work focuses on unlabeled data which are from another different source domain and a limited amount of target data. In other words, we do not target the semi-supervised learning setting.

Our work is complementary to recent self-supervised learning works \cite{caron2018deep,wu2018unsupervised,he2020momentum,chen2020simple,grill2020bootstrap,caron2020unsupervised} and studies how to leverage self-supervised pretraining for better few-shot learning. We observe that the new transductive self-supervised pretraining can significantly improve the performance of the vanilla self-supervised pretraining towards that of supervised pretraining in the few-shot learning.

\vspace{0.5em}
\noindent\textbf{Few-shot Learning.}
Few-shot learning aims to learn from a limited amount of labeled examples. To avoid overfitting caused by the data scarcity, most studies follow the paradigm of episodic-based learning, and can be divided into two main categories: meta-learning based and fine-tuning based. Meta-learning based methods can be further divided into “learning to optimize”, which tried to make the learning adapt to a new task with parameter updating~\cite{FinnAL17,Sachin2017,mishra2018a,rusu2018metalearning}, and “learning to compare”, which aimed to learn a good feature embedding and classified a new sample based on the nearest labeled instances in feature space~\cite{vinyals2016matching,snell2017prototypical,sung2018learning,wan2019transductive}. Fine-tuning based methods~\cite{chen2018a,Dhillon2020A} followed the standard transfer learning procedure of supervised pretraining on auxiliary labeled data and fine-tuning on the few target data. Chen \etal~\cite{chen2018a} showed that a simple fine-tuning based method was already competitive to sophisticated meta-learning based methods. Our work follows the fine-tuning based paradigm as well, but we study unsupervised pretraining using auxiliary unlabeled examples and propose a new sampling strategy to further boost the few-shot performance.

\vspace{0.5em}
\noindent\textbf{Unsupervised Few-shot Learning.} Recent work~\cite{antoniou2019assume,hsu2018unsupervised,khodadadeh2019unsupervised} proposed unsupervised few-shot learning to tackle the huge demand of a big labeled auxiliary dataset in supervised few-shot learning. They all followed the episode-based paradigm and attempted to construct episode-based few-shot tasks with unlabeled samples, \eg, Hsu \etal~\cite{hsu2018unsupervised} proposed to obtain pseudo labels for unlabeled samples via clustering and constructed episode-based few-shot tasks with pseudo labels. Khodadadeh \etal~\cite{khodadadeh2019unsupervised} and Antoniou \etal~\cite{antoniou2019assume} proposed to generate a pseudo query set through data augmentation of randomly sampled unlabeled examples. Different from these methods, our work deviates from episode-based few-shot paradigm and studies self-supervised pretraining for few-shot learning. We do not assume the unlabeled data follows the same distribution as the target few samples.

\vspace{0.5em}
\noindent\textbf{Active Learning.} Our eigen-finetuning approach shares a similar goal to classical active learning \cite{lewis1995sequential,roy2001toward,bilgic2009link,beluch2018power,gal2017deep} that uses a learning algorithm to actively query the user to label new data points with the desired outputs. There are two key difference. On the one hand, we aim at the pretraining for few-shot learning,  which involves two different domains at the same time, \ie, source and target domains. However, most active learning methods only consider the target domain. On the other hand, existing learning focus on how to get an optimal sampling strategy for better interactivity, which is often complicated. By contrast, we find that simple clustering-based sampling is good enough for few-shot learning. It is because the transfer really benefits from both the pretraining and finetuning, but not solely active learning.

\section{Method}

We study the problem of few-shot classification in the paradigm of pretraining and finetuning. In the recent big transfer (BiT) work~\cite{chen2020big}, the model is first pretrained on an auxiliary large scale labeled dataset $\mathcal{S}^\#=\{x^s_i, y^s_i\}_{i=1}^M$ by supervised learning, and then finetune the model on the small scale target dataset $\mathcal{T}=\{x_j\}^N_{j=1}$ with few labeled samples, where $N\ll M$. The pretrained feature representation significantly benefits few-shot learning. In this paper, we aim to reduce the demand for labels in both pretraining and finetuning. We ask the questions: 
 \begin{itemize}
     \item \textit{Can the auxiliary large scale dataset $\mathcal{S^\#}$ be \textbf{totally unlabeled}?}
     \item \textit{Can we use \textbf{fewer labels} in $\mathcal{T}$ to achieve comparable transfer performance?}
 \end{itemize}
 
\subsection{Clustering Matters to Few-shot Transfer.}

To resolve the first question, we are inspired by the recent success of self-supervised learning and try the vanilla unsupervised transfer, \ie, pretraining the model on the auxiliary large scale unlabeled dataset $\mathcal{S}=\{x^s_i\}_{i=1}^M$ using the state-of-the-art unsupervised learning method MoCoV2~\cite{chen2020improved}, and then finetuning on the target dataset $\mathcal{T}$. However, the transfer performance is not so good as the supervised transfer BiT~\cite{chen2020big}. To explore the reason, we compare unsupervised and supervised pretrainings in terms of the distribution of target samples $\{x_j\}^N_{j=1} \in \mathcal{T}$ in the feature space through t-SNE \cite{hinton2002stochastic}. The visualization is shown in~\Fref{fig:case}, where the target Pet37 dataset~\cite{parkhi2012cats} is used as an example. As we can see in~\Fref{fig:case} (a) and (b), the features obtained from the supervised pretraining model are better clustered than those obtained from the unsupervised pretraining model. Based on the observation, we make intuitive sense that \textit{the clustering quality matters to few-shot transfer}.

The hypothesis can be further elucidated. If the target features are well clustered after pretraining, it is much easier to learn a good classifier even though only a few labeled samples are available in the following finetuning. Moreover, we study the relationship between the clustering quality and few-shot performance on several different pretrained models: unsupervised pretraining ResNet-50 on ImageNet-1k, supervised pretraining ResNet-50 on ImageNet-100 and ImageNet-1k respectively. We use the BCubed Precision (Cluster Acc) as the metric of clustering quality~\cite{amigo2009comparison}. The results shown in \Tref{tab:cluster_abl} demonstrate that the few-shot transfer performance is highly correlated to the clustering accuracy.

\begin{table}[!t]
\small
  \begin{center}
  \setlength{\tabcolsep}{1.8mm}{
    \begin{tabular}{lccc|c}  
    \shline
               & Unsup-1k & Sup-100 & Sup-1k & Trans  \\
    \hline
    Cluster Acc &    47.72      &    12.82     &  67.44      &  61.69 \\
    5-shot Acc & 70.93 & 45.65 & 77.94 & 75.10 \\
    \shline
    \end{tabular} }
    \captionof{table}{The clustering accuracy and few-shot performance of different pretrained models on Pet37. ``Unsup, Sup, Trans" indicate vanilla unsupervised, supervised and transductive unsupervised pretraining respectively, and ``-1k/100" indicates ImageNet-1k/100.} 
    \label{tab:cluster_abl}
     \end{center}
     \vspace{-1em}
\end{table}

\subsection{Transductive Unsupervised Learning}

\noindent\textbf{Contrastive Learning.} 
Contrastive learning is widely used in the unsupervised learning literature. Given an image, two different augmentations are used. One augmented sample is regarded as query $\bm{q}$ and the other is regarded as the positive key $\bm{k}^+$. All augmented versions of other images are regarded as negative keys $\{\bm{k}_i^-\}$. Formally, the contrastive loss is defined as:
\begin{equation}
\label{eqn:contrastive}
    \mathcal{L} = - \text{log}\frac{\text{exp}(\bm{\bm{z}_{\bm{q}}}\cdot\bm{z}_{\bm{k}}^+ / \tau)}{\text{exp}(\bm{\bm{z}_{\bm{q}}}\cdot\bm{z}_{\bm{k}}^+/\tau) + \sum_{i=0}^{K-1} \text{exp}(\bm{\bm{z}_{\bm{q}}}\cdot\bm{z}_{\bm{k_i}}^-/\tau)},
\end{equation}
where $\bm{z}_{\bm{q}},\bm{z}_{\bm{k}}$ are the feature embedding by feeding $\bm{q},\bm{k}$ into two separated encoders. The query encoder and key encoder share the same architecture, \ie, the backbone of the target network followed by an extra MLP-based projection head, but have different weights. 

Essentially, the contrastive loss encourages similar images to have a close embedding distance and conversely, dissimilar images to have a far distance. This behavior is indeed the same as that of clustering. In other words, \textit{contrastive learning is really to learn how to cluster samples}.

\vspace{0.75em}
\noindent\textbf{Transductive Unsupervised learning.} Inspired by the connection between contrastive learning and clustering, one key observation of this paper is that the target samples can be better clustered in feature space if those are involved into the contrastive learning phase. Since the unlabeled target samples together with the unlabeled source samples are used in unsupervised learning, we refer to this way as \textit{``transductive unsupervised learning''}. By contrast, existing vanilla unsupervised learning that only utilizes the source data can be regarded as ``inductive''. The improved clustering  significantly boosts the few-shot learning performance in the target domain. In \Tref{tab:cluster_abl} shows our clustering accuracy and its corresponding few-shot performance on the target domain, and the feature visualization is also shown in \Fref{fig:case} (c).

\vspace{0.75em}
\noindent\textbf{Sample Re-balancing.} Empirically, we find naively mixing up $\mathcal{T}$ and $\mathcal{S}$ with the ratio $1:1$ in the pretraining does not work well. As described above, the amount of the target images in $\mathcal{T}$ is often much smaller than the auxiliary dataset $\mathcal{S}$ in real applications. It causes serious learning imbalance and makes transductive unsupervised pretraining degrade to the vanilla unsupervised pretraining. Instead, we propose a simple and effective sample re-balancing strategy which mitigates this problem by increasing the percentage of target samples in the mixture of target data $\mathcal{T}$ and source data $\mathcal{S}$. Besides, finding a proper percentage $p$ is necessary, since a too large or small percentage $p$ will both cause the degradation of performance.

\begin{algorithm}[t]
\small
 \caption{\small{Anchor-Constrained KMeans in $\kappa$-th evolution}}
 \renewcommand{\algorithmicrequire}{\textbf{Input:}}
 \renewcommand{\algorithmicensure}{\textbf{Output:}}
 \begin{algorithmic}[1]
 \Require Set of target features $\cF=\{f_i\}^N_{i=1}$ (estimated cluster label of $f_i$ is denoted as $f_i^{L}$). Number of new clusters $K$. Set of anchors $\cA=\{a_j\}^m_{j=1}$, Maximum iteration of KMeans $t_{max}$.
 \Ensure $K$ cluster centers $\{\mu_j\}_{j=1}^{m+K}$
 \State \textbf{-- Initialize Centers:}
 \State ${\mu_{j}^0}\xleftarrow[]{}{a_j}$, $j = 1, ..., m$; randomly initialize $\mu_{m+1}^0,...,\mu_{m+K}^0$.
  \For{$ t = 1,\dots, t_{max} $}
  \State \textbf{-- Assign Samples to Cluster}: 
  \For{$ i = 1,\dots, N $} 
    \State $f_i^{L}=\text{arg} ~ \underset{j}{\text{min}} ~ ||f_i-\mu_{j}||^2$, $j=1,...,m+K$
  \EndFor
  \State \textbf{-- Update Cluster Centers:}
  \State $\mu_j^{t}=\mu_j^{t-1}$, $j=1,...,m$
  \For{$ j = m+1,\dots, m+K $}
      \State $\cF_j^t=\{f_i|f_i^{L}=j, f_i\in \cF\}$,\vspace{0.3em}
      \State $\mu_j^{t}=\frac{1}{|\cF_j^t|}\sum_{f\in\cF_j^t}{f}$,
  \EndFor
  \EndFor
 \end{algorithmic}
 \label{alg:AC_KMeans}
\end{algorithm}

\subsection{Eigen-Finetuning for Fewer shot}
More interestingly, the improved clustering motivates us to address the second question, \ie, using fewer labels in the target dataset to achieve better results. The motivation is reflected in three perspectives: \vspace{-0.5em}
\begin{itemize}
\item The target samples closer to the clustering centers are more representative, which suggests choosing such samples to label can be more effective, especially under a very limited label budget. \vspace{-0.5em}
\item Finetuning the model with such labeled samples can further improve the clustering quality of all target samples, and in return, the improved model continues to help identify more representative samples.\vspace{-0.5em}
\item The clusters tend to be evolved from coarse to fine with the increase of labels, where the labeled representative samples at early evolution stand for coarse clusters and those at late evolution more likely act as fine clusters.  \vspace{-0.5em}
\end{itemize}

\vspace{0.75em}
\noindent\textbf{Eigen-Finetuing.} Integrated with the above aspects, a new \textit{eigen-finetuning} is proposed. We call the representative sample as \textit{eigen-sample}. As shown in \Fref{fig:pipeline}, the few-shot finetuning is converted into a co-evolution process: {``\textit{clustering} $\rightarrow$  \textit{eigen-samples annotation} $\rightarrow$ \textit{model finetuning} in a loopy way"}. In spirit, the way is similar to active learning. Specifically, at each evolution step $\kappa$, we first re-cluster the target features and incrementally find some eigen-samples, then annotate the new eigen-samples, and finally finetune the model with all the labeled eigen-samples. This co-evolution process will end until we reach the total annotation budget. By this way, the early found eigen-samples help collect target samples into coarse clusters, while the newly found eigen-samples will further help the relatively hard samples gather towards finer clusters. 

We develop a new clustering algorithm called \textbf{A}nchor \textbf{C}onstrained \textbf{KMeans} (\textbf{ACKMeans}) to implement the incremental eigen sampling. All eigen-samples found at previous $\kappa-1$ evolution steps, are referred to \textit{anchors}. The key idea of ACKMeans (at $\kappa$-th evolution) is that the anchors as cluster centers won't be changed during KMeans and help exclude samples close to these anchors; while the remaining of dissimilar samples would be clustered into $K$ new clusters, which helps select $K$ new eigen-samples to annotate. This way allows us to optimize the annotation budget to the most extent, since each eigen-sample represents a cluster of similar samples associated to it. At every evolution, supposing $b$ annotation budget per category, totally $K=b \times C$ new eigen-samples are chosen to be annotated, where $C$ denotes the number of target categories. Hence, the total annotation budget for target data would be $K \times \kappa_{max}$, where $\kappa_{max}$ is the maximum evolution steps.

\subsection{``$1+\epsilon$"-shot Setting for Few-shot}

The existing few-shot setting assumes all the labels are pre-known, and randomly chooses a certain percentage (\eg, $5\%$) of labeled samples per category to guarantee the few shots are class balanced. It can be regarded as an ideal or oracle few-shot setting, since consuming labels beforehand is usually unrealistic in real applications or it costs for the annotator to watch and label more samples beyond few shots. 

To address this issue, we recommend a more practical few shot setting: ``$1+\epsilon$"-shot. Initially, only ``$1$" image per target category is given. We think it reasonable since each category may need an indicator image when the annotation process starts. Next, ``$\epsilon$" extra annotations are required to be labeled for each category on average, and thus the total annotation cost $\epsilon \times C = K \times \kappa_{max}$. We do not guarantee each category can get exactly $\epsilon$ extra labels, but our sampling strategy achieves as possible as it can. Compared with the existing ideal few-shot setting, ``$1+\epsilon$"-shot tends to select $\epsilon$ most representative samples per category on average to label, in contrast to random sampling per category, which may unfortunately label very similar samples, being not optimal for the budget utilization. 

``$1+\epsilon$"-shot is suggested to real-world applications and makes it possible to use fewer labels to achieve better few-shot performance. In the experiments, we demonstrate our method can outperform both vanilla unsupervised transfer and supervised transfer even though they use the oracle few-shot setting.

\begin{table*}[!t]
\small
  \begin{center}
  \setlength{\tabcolsep}{1.2mm}{
    \begin{tabular}{lccccccccccc|c}  
    \toprule  
   1+ $\epsilon$ & Method 
   & DTD   & Food101 & SUN397   & Flower102  & Caltech101 & STL10  & CIFAR10  & CIFAR100  & EuroSAT & Pet37 & Mean Acc.\cr 
    \hline \hline
    \multirow{3}{*}{1+1}& MoCoV2 
    & 38.06 & 15.80    & 24.28   & 52.60    & 63.62      & 75.06   &  41.29   & 22.95     & 60.20  & 56.87 & 45.03 \cr
     & BiT 
    & 44.66 & 24.99   & 27.21   & 65.60    & 61.07      & 74.80   & 59.05    & \textbf{37.40}     & 68.31  & 63.95 &  52.70 \cr 
     & Ours 
    & \textbf{44.86} & \textbf{35.92}   & \textbf{35.46}   & \textbf{80.97}    & \textbf{79.87}   & \textbf{82.45}   & \textbf{62.19}    & 32.74     & \textbf{69.62}  & \textbf{68.87}  & \textbf{59.30} \cr 
    \hline
    \multirow{3}{*}{1+3} & MoCoV2  
    & 48.20  & 25.49  & 35.06   & 72.22   & 76.54       & 87.35    & 50.30    & 35.08     & 69.18 & 68.94 &  56.84 \cr
    & BiT 
    & 53.69  & 35.26  & 36.72   & 78.49   & 73.89      & 83.28     & \textbf{71.90}    & \textbf{47.79}  & \textbf{79.05} & \textbf{76.08} &  63.61\cr 
    & Ours 
    & \textbf{57.11}  & \textbf{49.66}  & \textbf{45.38}   & \textbf{89.81}   & \textbf{86.13}  & \textbf{88.19}     & 71.20    & 46.80    & 76.7   & 74.07 & \textbf{68.51} \cr 
    \hline
    \multirow{3}{*}{1+5} & MoCoV2 
    & 53.00   & 31.01  & 40.71   & 80.76   & 82.14      & 89.21     & 56.44    & 42.74    & 72.15  & 72.35 & 62.05 \cr
    & BiT 
    & 57.64  & 41.26  & 41.11   & 83.71   & 80.41      & 86.71     & \textbf{75.62}    & \textbf{54.17}  & \textbf{82.55}   & \textbf{79.85} &  68.30 \cr  
    & Ours 
    & \textbf{61.56}  & \textbf{56.50}  & \textbf{49.42}   & \textbf{92.82}   & \textbf{88.11}   & \textbf{90.14}     & 74.89    & 53.74   & 79.29   & 79.01 & \textbf{72.55}   \cr 
     \hline
     \multirow{3}{*}{1+7} & MoCoV2  
    & 56.00  & 35.50  & 44.85   & 84.91   & 84.45      & 90.21     & 57.87    & 47.84   & 74.36   & 75.12  & 65.11 \cr
    & BiT 
    & 61.04  & 45.60  & 43.22   & 86.90   & 83.87      & 88.91     & 78.10    & 57.46   & \textbf{84.71}   & \textbf{82.85}  &  71.22 \cr  
    & Ours 
    & \textbf{64.38}  & \textbf{60.15}  & \textbf{52.36}   & \textbf{93.97}   & \textbf{89.20}    & \textbf{90.98}   & \textbf{78.72}   & \textbf{57.86}   & 81.19   & 81.99  & \textbf{75.08}  \cr 
     \hline
    \multirow{3}{*}{1+9} & MoCoV2 
    & 58.68  & 38.38 & 47.75   & 87.42   & 86.75      & 90.85     & 59.92    & 51.87    & 75.69  & 77.62 &  67.49 \cr
    & BiT 
    & 63.06  & 48.77  & 44.96   & 88.58   & 86.29      & 90.07     & 79.99    & 59.92    & \textbf{86.25}  & \textbf{84.81}  & 73.27 \cr  
    & Ours 
    & \textbf{66.58}  & \textbf{62.67}  & \textbf{54.30}   & \textbf{94.62}   & \textbf{89.55}      & \textbf{92.02}     & \textbf{80.73}    & \textbf{60.63}    & 81.42   & 84.41  & \textbf{76.69} \cr 
    \bottomrule  
    \end{tabular} }
    \captionof{table}{\small  \textbf{The few-shot learning results on 10 benchmark target datasets.} Both the vanilla unsupervised pretraining baseline MoCoV2 and the supervised pretraining baseline BiT use the ideal few-shot learning setting, and our method uses the realistic and challenging $1+\epsilon$ setting.} 
    \label{tab:overall_results}
     \end{center}
    \vspace{-2em}
\end{table*}

\section{Experiments}
\subsection{Experimental Setup}
\noindent \textbf{Datasets.} In the following experiments, we use the ImageNet-1K dataset~\cite{deng2009imagenet} as the auxiliary large scale source dataset, and consider $10$ 
small target datasets: Pet37~\cite{parkhi2012cats}, SUN397~\cite{xiao2010sun}, DTD~\cite{cimpoi2014describing}, Flower102~\cite{nilsback2008automated}, CIFAR10 and CIFAR100~\cite{krizhevsky2009learning}, Caltech101~\cite{fei2004learning}, Food101~\cite{bossard2014food} and EuroSAT~\cite{helber2019eurosat}. These datasets are very diverse and differ
in the total image number, input resolution and nature of their categories, ranging from general object categories (\eg, CIFAR10/100) to fine-grained
ones (\eg, Pet37 and Flower102). We follow the standard setting as \cite{chen2020simple,grill2020bootstrap,kolesnikov2019big}, and report the mean class accuracy for Pet37, Flower102, Caltech101 and the Top1 accuracy for other datasets. All the results are averaged by 5 trials to reduce randomness.

\vspace{0.5em}
\noindent \textbf{Pretraining Details.} We build our transductive unsupervised learning based on MoCoV2~\cite{chen2020improved} and follow its training protocol. In details, we adopt the SGD optimizer with momentum $0.9$ and the weight decay $0.0001$. The initial learning rate is $0.24$ with a cosine scheduler and the batch size is 2,048. All the pretraining models are trained with 800 epochs. The backbone network for all the experiments uses ResNet-50~\cite{he2016deep}. The default sample re-balancing ratio varies based on the target dataset size so that the resampled target data size is about $20\%$ of the source dataset size. All the images are resized during training so that the short edge is 256 and randomly cropped into $224\times 224$.

\vspace{0.5em}
\noindent \textbf{Finetuning Details.} We finetune the pretrained model for $60$ epochs with a batch size of $256$, and the weight decay is $0$. The learning rate for the newly added \textit{FC} layer and pretrained layers is $3.0$ and $0.0001$ respectively. We only use random crop with resizing, flips for training and the center crop with resizing for testing. Particularly, for low-resolution datasets CIFAR10/100, STL10, EuroSAT, the resized short edge is $160$ and the crop size is $128$. For other datasets, we adopt the common $(256,224)$ combination. $b$ is set to be 1 for eigen-finetuing.

\subsection{Overall Results}

\Tref{tab:overall_results} reports the ``$1+\epsilon$''-shot results on all benchmark datasets. For comparison, we also report the results of vanilla unsupervised pretrained models (MoCoV2~\cite{chen2020improved}) and the supervised pretrained models (BiT~\cite{kolesnikov2019big}) under the ideal few-shot setting, which select ``$1+\epsilon$'' labeled samples for each category in a strictly class-balance way. Here, our method adopts the exactly same pretraining and finetuning setting to MoCoV2~\cite{chen2020improved}, and BiT directly uses the officially released code for pretraining and finetuning.

We can observe the following results. 1) Our method consistently outperforms the vanilla unsupervised pretraining baseline MoCoV2 across all the datasets by a large margin. Especially, on the Food101 dataset, the performance gain is consistently about $20$ points. 2) Our method outperforms the supervised pretraining baseline BiT on majority of datasets and is comparable or slightly worse on the rest. On average, our method performs better than BiT.

By analyzing the performance among different datasets, we further get some fine-grained observations:
\textit{1) 
Our method outperforms both vanilla unsupervised pretraining and supervised pretraining when the gap between source and target domains 
is either very large (\eg, SUN397) or very small (\eg. Caltech101)}.
For example, SUN397 is for scene recognition while ImageNet is almost object-centric. Therefore, either the supervised pretraining model or the vanilla unsupervised pretraining model cannot obtain good clustering on the target domain. (their Cluster ACC~\cite{amigo2009comparison}: $22.93\%$ \textit{vs.} $20.11\%$). In contrast, Caltech101 is object-centeric and shares similar categories with ImageNet, therefore both the supervised and the vanilla unsupervised pretrainings on ImageNet can achieve good clustering (their Cluster ACC: $47.11\%$ \textit{vs.} $53.14\%$). In both cases, the clustering quality of the vanilla unsupervised pretraining is much closer to that of supervised pretraining, compared with the results on Pet37 (see \Tref{tab:cluster_abl}). By involving the target data, our method can improve the clustering quality (Cluster ACC: $34.36\%$ on SUN397, $59.88\%$ on Caltech101) especially for large domain gap (SUN397), thus bringing significant performance gain. 

\textit{2) Though our method does not require a large target dataset, we empirically find it will bring more benefits if the target dataset has a larger scale.} One typical example is the Food101 dataset. It has a total of about 75k high-resolution images and each category has about 750 images. It is consistent with the common sense that bigger data can help learn better representation. A similar ablation study will be also given in the following parts.

\textit{3) Our method is comparable to or slightly worse than supervised pretraining if the target dataset has a low image resolution.} For example, the image resolution of CIFAR10/100 and EuroSAT is only $32\times 32$ and $64\times 64$, so directly upsampling them to match the image resolution on ImageNet may not be a good way for our method. In addition, STL10 has similar categories as CIFAR10 but a larger image resolution, and thus our method achieves better performance in STL10 than in CIFAR10.

\subsection{More Analysis about Unsupervised Pretraining}

\begin{figure}[!t]
	\centering
	\includegraphics[width=1.0\linewidth]{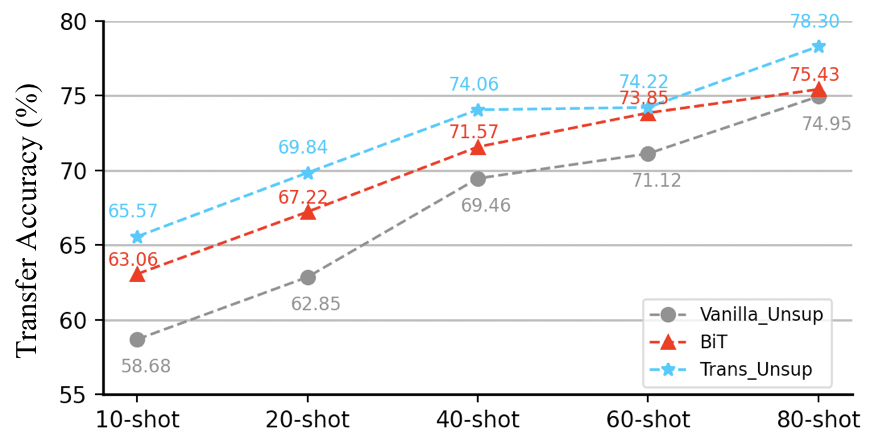}
	\caption{\small \textbf{Transductive unsupervised learning vs. Vanilla unsupervised learning vs. Supervised learning}. The evaluation is performed by using a different number of labeled samples for finetuning on the DTD dataset. The ideal few-shot setting is used.}
	\label{fig:label_vary_2}
\end{figure}

\begin{table}[t]
\small
  \begin{center}
  \setlength{\tabcolsep}{3.3mm}{
    \begin{tabular}{ccccc}  
    \toprule  
   {Dataset}& Method & 1-shot & 5-shot & 10-shot \cr  
    \hline \hline
     \multirow{2}{*}{DTD}
     & Vanilla  & 26.74             & 50.54             & 58.68  \cr
     & Trans    & \textbf{32.07}    & \textbf{58.01}    & \textbf{65.57}  \cr 
      \hline
     \multirow{2}{*}{Food101}
     & Vanilla  & 10.09             & 28.39             & 38.38   \cr
     & Trans    & \textbf{23.55}    & \textbf{51.99}    & \textbf{62.56}  \cr
     \hline
     \multirow{2}{*}{CIFAR100}
     & Vanilla  & 13.65             & 39.53             & 51.87   \cr
     & Trans    & \textbf{18.82}    & \textbf{49.08}    & \textbf{60.91}  \cr
    \bottomrule  
    \end{tabular} }
    \captionof{table}{\small \textbf{Transductive unsupervised learning vs. Vanilla unsupervised learning}. The accuracy is evaluated under the classical ideal few-shot setting.} 
    \label{tab:trans_pretrain}
     \end{center}
    \vspace{-1em}
\end{table}
In this section, we continue to compare the transfer performance between unsupervised pretraining and supervised pretraining in depth, for not only few-shot learning but also learning on the whole dataset. Specifically, we take the DTD dataset (80 labeled samples per category) as an example, and test the transfer performance of different pretrained models by varying the number of labeled samples during finetuning. As shown in Figure \ref{fig:label_vary_2}, although supervised pretraining is superior to the vanilla unsupervised pretraining by a considerable margin, the performance gap becomes much smaller as the number of labeled samples is increasing. A similar trend can also be observed in other datasets.

Therefore, we have the following observations. 1) For few-shot transfer, the vanilla unsupervised pretraining is often inferior to supervised pretraining. This should be related to the key insight that unsupervised learned representation is not compact as the supervised counterpart and thus results in worse clustering. 2) The representation from the unsupervised pretraining model itself is not bad. Given a moderate number of labeled samples, it can match or even beat the transfer performance of the supervised counterpart. This is consistent with the conclusion drawn in some existing unsupervised learning work~\cite{chen2020simple,grill2020bootstrap}. 3) By involving the target dataset into the pretraining, our proposed transductive unsupervised pretraining can compensate the clustering ability of features from unsupervised pretraining. Therefore, our method can achieve better results for both few-shot learning and full-dataset transfer.

\subsection{Ablation Study}
In this section, extensive ablation studies are conducted to analyze different components of our method. 

\vspace{0.5em}
\noindent  \textbf{Benefits from Transductive Unsupervised Pretraining.}
In this experiment, we adopt the typical ideal few-shot setting to show the advantage of transductive unsupervised pretraining over the vanilla unsupervised pretraining. Because of the space limit, we only list the results of three representative datasets in Table \ref{tab:trans_pretrain}. As we can see, it yields consistent performance gain upon the vanilla baseline and makes it possible to use fewer labels for few-shot learning. For example, the 5-shot performance by transductive unsupervised pretraining is competitive and even better than the 10-shot performance by the vanilla counterpart.

\begin{figure}[!t]
	\centering
	\includegraphics[width=1.0\linewidth]{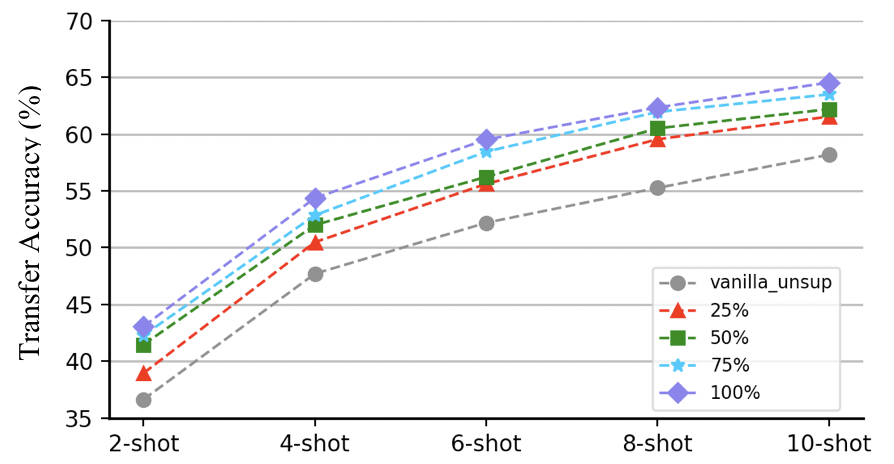}
	\caption{\small \textbf{Different percentages of unlabeled target data} involved into transductive unsupervised pretraining. The few-shot performance is used for evaluation. Our method can benefit from using more unlabeled data in the target domain.}
	\label{fig:sample_ratio}
\end{figure}

\begin{table}[t]
\small
  \begin{center}
  \setlength{\tabcolsep}{2.3mm}{
    \begin{tabular}{cccccc}  
    \toprule  
      balance ratio & 1+1 & 1+3 &  1+5 & 1+7  &  1+9 \cr  
    \hline \hline
    w/o re-balance  & 40.27       &  49.55           & 54.42          &  59.28            & 61.01 \cr
      20\%        & \textbf{44.86} &  \textbf{57.11}  & \textbf{61.56} &  \textbf{64.38}   & \textbf{66.58}  \cr
      50\%        & 38.83          &  50.06           & 55.27          &  60.59            & 63.86  \cr 
    \bottomrule  
    \end{tabular} }
    \captionof{table}{\small \textbf{Different sampling re-balancing ratios during pretraining} evaluated by using the few-shot performance on the DTD dataset.} \label{tab:balance_ratio}
     \end{center}
    \vspace{-1em}
\end{table}

\vspace{0.5em}
\noindent  \textbf{Influence of Target Dataset Scale in Pretraining.} To verify the hypothesis that our method will benefit from a larger amount of unlabelled target dataset, we further conduct a simple ablation experiment on the DTD dataset. Specifically, during the transductive unsupervised pretraining stage, we involve different percentages of target data ($25\%, 50\%, 75\%, 100\%$), and then evaluate the transfer performance under the ideal few-shot setting. As shown in \Fref{fig:sample_ratio}, involving more unlabeled target data into pretraining can help learn better representation, thus producing better few-shot performance. It further shows the merit of our transductive unsupervised pretraining from two aspects. 1) Since it is often much easier and cheaper to collect unlabeled target dataset than labeled target dataset, our pretraining setting is more scalable than supervised pretraining. 2) For the cases where a big unlabeled target dataset exists, our method can fully utilize the power of big data while the vanilla unsupervised pretraining may not.

\vspace{0.5em}
\noindent \textbf{Ablation of Sample Re-balancing Ratio.} As described in the method part, the target datasets often have a small image amount in some real applications, which can be smaller than auxiliary source dataset by several magnitudes. Therefore, we find sample re-balancing is indispensable to relieve the data imbalance issue during pretraining. Here, we use the DTD datast as an example and try two variants: without sample re-balancing and with a large re-balancing ratio (resampled target dataset size is $50\%$ of the source dataset size). As we can see in \Tref{tab:balance_ratio}, the transfer performance degrades if no sample re-balancing is applied, and too large re-balancing ratio will also lead to inferior results because the benefit from the auxiliary source dataset is suppressed. Therefore, we empirically use a re-balancing ratio $\sim 20\%$ in all the experiments.

\begin{table}[t]
\small
  \begin{center}
  \setlength{\tabcolsep}{1.68mm}{
    \begin{tabular}{ccccccc}  
    \toprule  
   {Dataset}&  & 1+1 & 1+3 &  1+5 & 1+7  &  1+9 \cr  
    \hline \hline
     \multirow{3}{*}{DTD}
     & Random  & 40.15          &  51.19         & 56.72          &  60.75          & 63.48 \cr
     & Oracle  & 44.79          &  55.17         & 60.05          &  63.04          & 65.57  \cr
     & Ours    & \textbf{44.86} & \textbf{57.11} & \textbf{61.56} & \textbf{64.38}  &\textbf{66.58}  \cr 
      \hline
     \multirow{3}{*}{Food101}
     & Random  & 32.03          &  46.36         & 54.50          &   58.48         &  62.00  \cr
     & Oracle  & 34.27          & 48.03          & 55.35          &   59.79         &  62.56  \cr
     & Ours    & \textbf{35.92} & \textbf{49.66}  & \textbf{56.50} & \textbf{60.15}   & \textbf{62.67}  \cr 
      \hline
     \multirow{3}{*}{CIFAR100}
     & Random  & 26.82          &  41.63         & 50.19          &   55.82         &  58.99  \cr
     & Oracle  & 30.24          &  43.94          & 52.68          &   57.73         &  \textbf{60.91}  \cr
     & Ours    & \textbf{32.74} & \textbf{46.80}  & \textbf{53.74} & \textbf{57.86}   & 60.63  \cr 
    \bottomrule  
    \end{tabular} }
    \captionof{table}{\small \textbf{Our proposed sampling strategy vs. Random sampling vs. Oracle-based sampling}. For all experiments, transductive unsupervised pretrained models are used. } \label{tab:pro_sample}
     \end{center}
    \vspace{-1em}
\end{table}

\noindent  \textbf{Effectiveness of Eigen-finetuing.}
To investigate the effectiveness of eigen-finetuing, we compare our sampling strategy with two baseline sampling strategies: 1) \textit{random sampling strategy}, which randomly selects the pre-defined number of samples from unlabeled target set and cannot guarantee class-balance; 2) \textit{oracle based sampling strategy} like ideal few-shot setting, which assumes all the labels pre-known and randomly samples an equivalent number of images across various categories in a class balance way. As shown in \Tref{tab:pro_sample}, the progressively-clustering-based sampling used in eigen-finetuing is much better than the random sampling, and competes or even outperforms the oracle based sampling strategy. On the DTD dataset, our ``$1+7$"-shot result can match the ``$1+9$''-shot result of random sampling, making fewer labels possible in real applications. 

\begin{figure}[!t]
	\centering
	\includegraphics[width=1.0\linewidth]{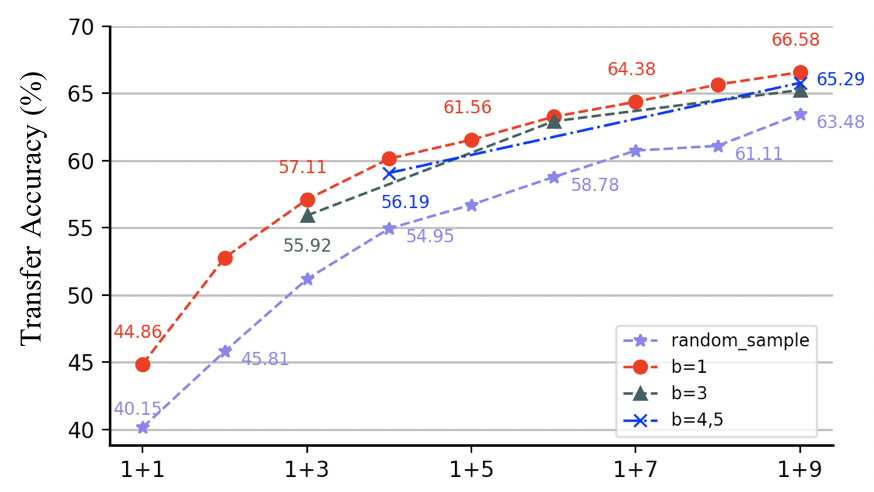}
	\caption{\small \textbf{The number of selected eigen-samples} to query labeling at each eigen-finetuning step. } \label{fig:b_study}
\end{figure}

\noindent \textbf{Ablation of the Annotation Number $b$.} In our default implementation of eigen-finetuning, we set $b=1$ at each evolution step of eigen-finetuning. However, for real applications, we can also set $b>1$ to reduce the total evolution step number and annotate more images at each evolution step. To demonstrate the generalization ability of eigen-finetuning with different $b$ values, we design two simple ablation experiments. In details, suppose the maximum annotation budget is $10 \times C$, we try two different finetuning strategies on the DTD dataset, namely, we either finish eigen-finetuning with 3 steps by setting $b=3$ for each step, or with 2 steps by setting $b=4$ for the first step and $b=5$ for the second step. As shown in \Fref{fig:b_study}, these two coarse strategies achieve slightly worse performance than the default finegrained strategy ($b=1$), but still outperform the random sampling baseline by a large margin. 
In the real application scenarios, by setting $b$ different values, our method can provide the flexibility to achieve a trade-off between performance and training efficiency.

\begin{table*}[t]
\small
  \begin{center}
  \setlength{\tabcolsep}{1.2mm}{
    \begin{tabular}{lccccccccccc}  
    \toprule  
   1+ $\epsilon$ & Method 
   & DTD   & Food101 & SUN397   & Flower102  & Caltech101 & STL10  & CIFAR10  & CIFAR100  & EuroSAT & Pet37 \cr 
    \hline \hline
    \multirow{5}{*}{1+1}& MoCoV2 
    & 38.06 & 15.80    & 24.28   & 52.60    & 63.62      & 75.06   &  41.29   & 22.95     & 60.20  & 56.87  \cr
     & BiT 
    & 44.66 & 24.99   & 27.21   & 65.60    & 61.07      & 74.80    & \textbf{59.05}    & \textbf{37.40}     & \textbf{68.31}  & 63.95  \cr 
    & Ours-2 
    & -      &   -      & \textbf{32.94}      &     -     & 79.26  & -                & -                & -        & -         & -   \cr 
    & Ours-5 
    & 47.02  & \textbf{37.03}   & 31.66       & 91.29     & 78.43  & -                & -                & -        & -         & -   \cr 
     & Ours-10 
    & \textbf{48.24} & 35.61   & 31.31        & \textbf{91.65} & \textbf{79.80}   & \textbf{81.32}   & 49.12    & 27.86     & 66.79  & \textbf{68.20}   \cr 
    \hline
    \multirow{5}{*}{1+3} & MoCoV2  
    & 48.20  & 25.49  & 35.06   & 72.22   & 76.54       & 87.35    & 50.30    & 35.08     & 69.18 & 68.94  \cr
    & BiT 
    & 53.69  & 35.26  & 36.72   & 78.49   & 73.89      & 83.28     & \textbf{71.90}    & \textbf{47.79}  & \textbf{79.05} & 76.08 \cr 
    & Ours-2 
    & -               &  -              & \textbf{43.40}   &    -             & 85.14           & -                & -         & -        & -       & -  \cr
    & Ours-5 
    & 57.41           & \textbf{52.11}  & 41.67            & \textbf{94.44}   & 84.88           & -                & -         & -        & -       & -  \cr
    & Ours-10 
    & \textbf{58.34}  & 48.73           & 40.73            & 94.18            & \textbf{85.52}  & \textbf{88.71}    & 59.63    & 41.37    & 76.59   & \textbf{76.87}  \cr 
    \hline
    \multirow{5}{*}{1+5} & MoCoV2 
    & 53.00   & 31.01  & 40.71   & 80.76   & 82.14      & \textbf{89.21}     & 56.44    & 42.74    & 72.15  & 72.35  \cr
    & BiT 
    & 57.64  & 41.26  & 41.11   & 83.71   & 80.41       & 86.71     & \textbf{75.62}    & \textbf{54.17}  & \textbf{82.55}   & \textbf{79.85}  \cr
    & Ours-2 
    &     -           & -               & \textbf{48.03}   &          -      &     \textbf{87.81}    &     -              & -        & -       & -       & -    \cr
    & Ours-5 
    & 60.36           & \textbf{57.72}  &          46.69   & \textbf{95.33}   &     86.56    &     -              & -        & -       & -       & -    \cr
    & Ours-10 
    & \textbf{61.25}  &   55.20         &     45.46        &   94.64          &     87.65    & 88.94     & 66.90    & 48.51   & 80.66   & 79.34    \cr
     \hline
     \multirow{5}{*}{1+7} & MoCoV2  
    & 56.00  & 35.50  & 44.85   & 84.91   & 84.45      & 90.21     & 57.87    & 47.84   & 74.36   & 75.12   \cr
    & BiT 
    & 61.04  & 45.60  & 43.22   & 86.90   & 83.87      & 88.91     & \textbf{78.10}    & \textbf{57.46}   & \textbf{84.71}   & \textbf{82.85}   \cr 
    & Ours-2 
    & -               & -               & \textbf{50.84}            & -                &     88.30         & -           & -   & -  & -   & -   \cr 
    & Ours-5 
    & \textbf{63.13}  & \textbf{60.91}  & 49.57   & \textbf{95.70}   & \textbf{88.96}    & -           & -   & -  & -   & -   \cr 
    & Ours-10 
    &        62.83   &    58.74  &     48.16        &     95.58        & 88.64    & \textbf{91.24}   & 69.55   & 53.82   & 81.22   & 80.05   \cr 
     \hline
    \multirow{5}{*}{1+9} & MoCoV2 
    & 58.68  & 38.38 & 47.75   & 87.42   & 86.75      & 90.85     & 59.92    & 51.87    & 75.69  & 77.62  \cr
    & BiT 
    & 63.06  & 48.77  & 44.96   & 88.58   & 86.29      & 90.07     & \textbf{79.99}    & \textbf{59.92}    & \textbf{86.25}  & \textbf{84.81}   \cr 
    & Ours-2 
    &    -       & -        & \textbf{53.30}      &   -              & \textbf{89.54}      & -             & -       & -        & -   & -  \cr 
    & Ours-5 
    &    64.53   & \textbf{63.13}       & 51.49   & \textbf{96.02}   &     89.22           & -             & -       & -        & -   & -  \cr 
    & Ours-10 
    & \textbf{64.73}  &     60.88       &    50.54         &         95.59    &          89.06      & \textbf{92.29}     & 72.04    & 56.29    & 84.76   & 81.44   \cr 
    \bottomrule  
    \end{tabular} }
    \captionof{table}{\small \textbf{Transductive unsupervised pretraining involving two target datasets(Ours-2), five target datasets(Ours-5) and ten target datasets(Ours-10) simultaneously} . Both the vanilla unsupervised pretraining baseline MoCoV2 and the supervised pretraining baseline BiT use the ideal few-shot learning setting, and our method uses the challenging $1 + \epsilon $ setting.}  
    \label{tab:multi_dataset}
     \end{center}
    \vspace{-2em}
\end{table*}


\noindent \textbf{Pretraining on Multiple Target Datasets.} 
In this ablation, we study the effectiveness of transductive unsupervised pretraining on multiple target datasets by involving two, three, or ten target datasets into pretraining all at once. Compared to our default setting that involves each target dataset independently, this setting is more efficient. The detailed comparison results are shown in Table \ref{tab:multi_dataset}, where ``-" means the target datasets not involved. It can be seen that transductive unsupervised pretraining involving multiple target datasets can induce slight performance drop, yet it can still achieve substantial improvements over vanilla unsupervised pretraining and outperform the supervised pretraining on the majority of datasets. 
\section{Conclusion}
In this paper, we study a challenging few-shot learning problem, that is, using fewer labels to achieve better few-shot performance.  Our pretraining requires no labels and our finetuning can also utilize fewer labels for better learning. For this purpose, we propose a novel approach consisting of transductive unsupervised pretraining and eigen-finetuning. All the designs are based on the key insight that clustering on the target domain is really related to the few-shot learning performance. Experimental results on various target datasets demonstrate the advantage of our method over both the vanilla unsupervised pretraining and the supervised pretraining. We believe such a practical few-shot setup will become a good attempt to solve other real applications of computer vision.

{\small
\bibliographystyle{ieee_fullname}
\bibliography{egbib}
}

\end{document}